\documentclass[11pt,a4paper]{article}
\usepackage[hyperref]{eacl2021}
\usepackage{times}
\usepackage{latexsym}

\usepackage{xspace}
\usepackage{xargs}
\usepackage{xcolor}
\usepackage{stfloats}

\usepackage{marginnote}
\usepackage{booktabs}
\usepackage{amsmath}
\usepackage{amssymb}
\usepackage{hyperref}
\usepackage{appendix}
\usepackage{multirow}
\usepackage{tabularx,booktabs}
\usepackage{arydshln}
\usepackage{multirow}

\usepackage{url}
\usepackage{framed}
\definecolor{shadecolor}{RGB}{150,150,150}
\definecolor{mygreen}{rgb}{0.13, 0.55, 0.13}

\usepackage{microtype}
\usepackage[final]{changes}

\aclfinalcopy 

\newcommand{\ignore}[1]{}

\newcommand{\eat}[1]{\ignorespaces}

\title{Few-shot Mining of Naturally Occurring Inputs and Outputs}

\author{Mandar Joshi$^{\dagger}$ \quad Terra Blevins$^{\dagger}$ \quad Mike Lewis$^{\mathsection}$ \\  { \bf Daniel S. Weld$^{\dagger\epsilon}$ \quad Luke Zettlemoyer$^{\dagger\mathsection}$ } \\[4pt]
$^{\dagger}$ Allen School of Computer Science \& Engineering, University of Washington, Seattle, WA \\
{\tt \{mandar90,blvns,weld,lsz\}@cs.washington.edu}\\[4pt]
$^{\epsilon}$Allen Institute of Artificial Intelligence, Seattle\\
{\tt \{danw\}@allenai.org} \\[4pt]
$^{\mathsection}$ Facebook AI Research, Seattle\\
{\tt \{mikelewis,lsz\}@fb.com}
}

\date{}

\begin{document}
\maketitle

 \begin{abstract} 
Creating labeled natural language training data is expensive and requires significant human effort.
We mine input output examples from large corpora using a supervised mining function trained using a small seed set of only 100 examples. The mining consists of two stages -- (1) a biencoder-based recall-oriented dense search which pairs inputs with potential outputs, and (2) a crossencoder-based filter which re-ranks the output of the biencoder stage for better precision. Unlike model-generated data augmentation, our method mines \emph{naturally occurring} high-quality input output pairs to mimic the style of the seed set for multiple tasks. On SQuAD-style reading comprehension, augmenting the seed set with the mined data results in an improvement of 13 F1 over a BART-large baseline fine-tuned only on the seed set. Likewise, we see improvements of 1.46 ROUGE-L on Xsum abstractive summarization.    

 \end{abstract}
 
 \section{Introduction}

 Gathering high-quality labeled training data  has been one of the most reliable ways of achieving empirical progress for a range of natural language processing tasks. However, creating natural language training data often involves significant human effort -- in carefully designing the data collection tasks as well as getting contributors to perform them. The complexity of this process in turn drives up the cost of creating training data. In this paper, we mine naturally occurring  examples from large corpora using supervision from a small seed set of only 100 labeled examples.

 \begin{figure}[t]
\begin{tabular}{p{7cm}}

\hline
\textbf{Summary}: The Scottish city of Edinburgh is looking to crack down on so-called "silent disco" walking tours as residents complain they make too much noise. \\
\textbf{Passage}: Silent disco tours in Edinburgh are 'too loud' and could face clampdown Silent discos in Edinburgh could be facing a clampdown - with locals upset that the crowds of boogying tourists are too loud. The "discos" involve people wearing their own headphones and dancing along as they follow a guided walking tour of the Scottish capital's most famous spots. ... \\
\hline
\textbf{Question}: How long does it take for a crab to get full grown? \\
\textbf{Passage}: 10-13 moults the crab will reach maturity. This usually takes \emph{3-4 years}, but when food is limited it can take longer to reach maturity. \vspace{1pt}\\
\hline
\end{tabular}
    \caption{Examples of mined input output pairs for summarization and reading comprehension. Answer spans are indicated via italics.}
    \label{fig:intro_ex}
\end{figure}
 
 Our method provides a way to collect large amounts of training data using a small set of labeled seed examples, and allows for more direct control over what the model learns compared to relatively  brittle prompts~\cite{Lu2021FantasticallyOP}. Yet, unlike model generated data augmentation, we mine high-quality human-authored data which is less susceptible to the limitations of synthetic data (Section~\ref{sec:ablations}). While similar unsupervised methods have been used for mining parallel data for machine translation~\cite{Tran2020CrosslingualRF}, we show that they can be extended to other tasks using minimal human supervision.

 One of the main challenges in mining high-quality data is the large search space which could be quadratic in the size of the corpora as inputs and outputs could be spread across multiple documents.
 Our approach consists of a two-stage pipeline which first mines data efficiently from large corpora and then filters out low-quality examples. More specifically, we use supervision from an initial seed set and train a dot-product similarity function over separately encoded fixed length dense representations of inputs and outputs (i.e. a dual-encoder or a biencoder). This function efficiently searches top-k candidate outputs (e.g. summary sentences) for each candidate input (e.g. news document) from the output corpus. We refer to this stage as biencoder-based search. It relies on maximal inner product search (MIPS) over dense encodings on inputs and outputs for efficiency. The initial list of paired inputs and outputs from this stage is geared towards recall and often contains examples with subtle errors. For example, it might link a document about World Cup 2006 with a summary sentence about World Cup 2010.
 
For the second stage, we train a precision-oriented crossencoder-based filter which re-ranks the output of the first stage for better quality control. The crossencoder is trained using positive samples from the original seed set and negative examples from the biencoder-based stage. Jointly encoding inputs and outputs provides fine grained interaction between them. The crossencoder is able to filter out more subtle errors such as the one described earlier to give a high quality mined dataset set.
 
 We apply our method to reading comprehension and summarization, gathering high quality question-answer-passage and document-summary tuples respectively. The supervision from the small seed set helps embed relevant outputs closer to inputs across corpora for each task. This is in contrast to MT where cross-lingual cosine similarity functions (over mean pooled output representations) are readily available without additional human supervision -- since self-supervised pre-training embeds sentences across languages in the same space. 
 
 \begin{figure*}[!t]
\centering
\includegraphics[width=1\textwidth, keepaspectratio, clip ]{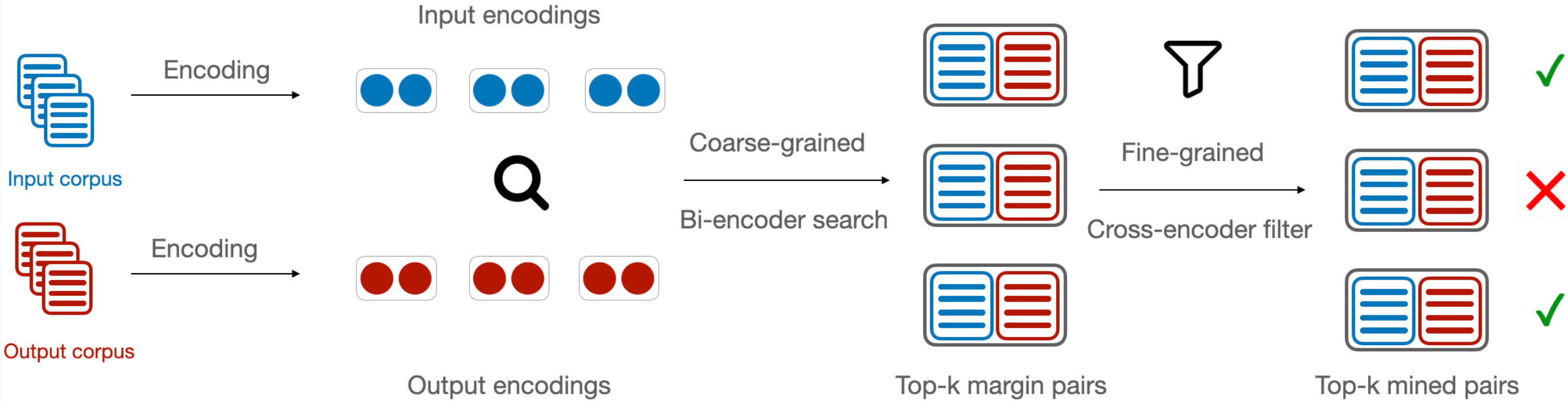}
\caption{A pipeline of bi- and cross-encoder models for mining high-quality data from large corpora. The bincoder search encodes the input and output corpora separately, and produces input-output pairs. The crossencoder filter then re-ranks these pairs to produce high-quality  data.}
\label{fig:pipeline}
\end{figure*}
 
 Our method is able to add up to 5x as many examples as the seed set.
 On SQuAD~\cite{rajpurkar2016squad}, augmenting the seed set with the mined data results in an improvement of 13 F1 over a BART-large~\cite{lewis-etal-2020-bart} baseline fine-tuned only on the seed set. Likewise, we see improvements of 1.46 ROUGE-L on XSum abstractive summarization~\cite{narayan-etal-2018-dont}. Our analysis shows that, compared to model generated data, mined data better matches several characteristics of the gold data. For example, mined document-summary pairs are more abstractive and are topically closer to XSum. Overall, mined data augmentation is based on the  intuition that learning similarity functions between contexts may be an easier problem that generating one context from another~\cite{khandelwal20generalization}. 
 
 \section{Mining}
 We describe our general pipeline for mining examples, and then describe how it is applied to reading comprehension and summarization. We assume access to a small labeled seed set of 100 examples which can be used to train components of the pipeline. The final output is a (ranked) list of input output pairs which can be augmented to the original seed set to train better-performing models. 
 
 The key idea consists of a two stage approach for mining input output pairs $(x,y)$ from their respective corpora $C_x$ and $C_y$ -- (a) a biencoder-based efficient similarity function which retrieves top-k similar candidates from $C_y$ for every $x \in C_x$ and (b) a crossencoder-based filtering stage which reranks top scoring candidate pairs mined in the first stage. 
 
 \subsection{Biencoder Search}
 The biencoder search relies on dense retrieval to construct a candidate list of input-output pairs. We first train a function $f_c(\cdot, \cdot)$ to maximize similarity between input output pairs in the seed data, and then use the similarity function to mine the candidate input output pairs.
 
 \subsubsection{Training Biencoder Similarity}
 \label{sec:sim}
 The aim is to create a vector space where the similarity between relevant input output pairs is higher than that between irrelevant pairs. Specifically, we use cosine similarity as our similarity function to enable efficient mining (Section~\ref{sec:mining}). 
 \paragraph{Objective}
 Let $D = \{(x_i, y_i^+, y_{i, 1}^-... y_{i, n}^-)\}_{i=1}^{i=m}$ be the training data that consists of $m$ seed instances. Each
instance contains one input $x_i$ and one output $y_i^+$ along with $n$ sampled negative outputs $y_{i, *}^-$. Our biencoder first encodes inputs and outputs separately as $\mathbf{x}_*$ and $\mathbf{y}_*$. The training objective maximizes the negative log likelihood of the positive output.

\begin{equation*}
    L(x_i, y_i^+, y_{i, 1}^-... y_{i, n}^-) = \frac{e^{\mathbf{x}_i \cdot \mathbf{y}_i^+}}{e^{\mathbf{x}_i \cdot \mathbf{y}_i^+} + \sum_{j=1}^n e^{\mathbf{x}_i \cdot \mathbf{y}_{i, j}^-}}
\end{equation*}
We discuss details about negative samples in Section~\ref{sec:application}.
 
 \subsubsection{Mining Candidate Pairs}
 \label{sec:mining}
 To mine input output pairs, we follow previous work on unsupervised machine translation and employ
the margin function formulation~\cite{artetxe-schwenk-2019-margin,Tran2020CrosslingualRF} based on K nearest neighbors (KNN). Let $\mathbf{x}$ and $\mathbf{y}$ be the vector representations of a candidate input output pair $(x, y)$. We score the similarity of $x$ and $y$ using a ratio margin
function defined as the following.
\begin{equation*}
    score(x, y) = \frac{cos(\mathbf{x},\mathbf{y})}{\sum_{z \in N_x} \frac{cos(\mathbf{x}, \mathbf{z})}{2k} + \sum_{z \in N_y} \frac{cos(\mathbf{z}, \mathbf{y})}{2k}}
\end{equation*}
 where $N_x$ is the KNN neighborhood of $x$ in $C_y$, the corpus of $y$; and $N_y$ is the
KNN neighborhood of $y$ in $C_x$. The margin scoring function is
interpreted as a cosine score normalized by average distances to the margin regions
established by the KNN neighborhoods of the input and outputs. The KNN distance metrics are defined by $cos(x, y)$. The margin score is designed to take into account scale inconsistencies in cosine similarity, and has been shown to perform better than approaches which use
a hard threshold over cosine similarity. We use the distributed dense vector
similarity search library FAISS~\cite{johnson2019billion} to search for all neighborhoods efficiently.
 
 \subsection{Crossencoder Filtering}
 We found the quality of the biencoder-mined data to be rather low for summarization and reading comprehension. Top ranked pairs for summarization were mostly extractive and those for reading comprehension often had very short passages (Section~\ref{sec:discussion}). We hypothesize that separate encoding of inputs and outputs combined with cosine based scoring makes it much harder to control  characteristics like abstraction for summarization.
 The crossencoder filtering stage re-ranks the outputs of the previous stage using a crossencoder based ranking function to remove noisy  input-output pairs. A crossencoder jointly encodes the input and output text enabling more fine grained interaction between them. Specifically, we encode a pair $(x, y)$ as sequence $[CLS]$ $x$ $[SEP]$ $y$ $[SEP]$ where $[CLS]$ and $[SEP]$ are special tokens indicating beginning and end of sequences.  The same input is fed into the encoder and decoder.
 The pair is scored using a 2-layer MLP on top of the  final hidden state of the final decoder token. A more detailed comparison of the data obtained from the two stages is presented in Section~\ref{sec:discussion}.

 \section{Application to Tasks}
 \label{sec:application}
 One of the main challenges in mining data is the scale of web corpora. In this section, we describe additional techniques which make our method scalable when applied to downstream tasks.
 
 \subsection{Reading Comprehension}
 \paragraph{Inputs and Outputs}
 For reading comprehension, the input sequence $X$ consists of the question and the output sequence $Y$ consists of a concatenation of the answer and the passage. Our input corpus consists of 20M random questions crawled from the popular community QA website \emph{answers.com}.  The output corpus consists of passages from Wikipedia with named entities\footnote{We use spaCy NER.} as candidate answers resulting in 200M passage-answer pairs. Our  input and output corpora were chosen to match the downstream evaluation data from SQuAD. 
 
 \paragraph{Biencoder and  Binary Classifier} To mitigate noise in the question corpus and to reduce the search space, we multitask a binary classifier with the biencoder to filter out questions which don't look like those in SQuAD. The binary classifier is trained using questions from the seed set as positive examples with negatives sampled from the question corpus. Furthermore, for faster processing, we filter out examples from the biencoder stage, where the answer spans are found verbatim in the question. 
 
 Apart from in-batch negatives, we also add synthetically generated negative samples for each question by (a) pairing incorrect candidate answer spans (identified using NER) with correct passages, and (b) pairing correct answers with incorrect passages. The aim of adding synthetic negatives is to avoid overfitting to passage-only and answer-only features.
 
 \paragraph{Crossencoder}
 The crossencoder for SQuAD is a simple binary classifier which uses the seed set as positive examples and the negative samples from the bicencoder as negatives.

 \subsection{Summarization}
 \paragraph{Inputs and Outputs}
 For summarization, the input sequence $X$ consists of the document and the output sequence $Y$ consists of the summary. The input corpus consists of documents from CC-News; we remove documents with less than four sentences to reduce noise. The output corpus consists of sentences from the same corpus. Following MARGE~\cite{Lewis2020PretrainingVP}, we divide CC-News into shards based on document publishing dates, and restrict the space of candidate summaries for a given document to be from the same shard. 
 
 \paragraph{Biencoder and Binary Classifier} Like reading comprehension, we multitask a binary classifier with the biencoder to remove sentences which don't resemble XSum summaries. This allows us to filter out more than 80\% of the sentences and makes the biencoder based mining a lot more scalable. 
 
 Finally, for faster processing, we filter out examples from the biencoder stage, where the summary sentences are found verbatim in the document. Put together, this allows use to efficiently search over 40M documents and 550M sentences. 
  We use in-batch summary sentences along with randomly sampled sentences from the output corpus to construct negative summary samples. 
  
  \paragraph{Crossencoder}
  Unlike reading comprehension, summarization has a larger space of \emph{valid} outputs. For example, it is possible to alter the style of the given summary without changing the core content. We tweaked the crossencoder training objective for summarization to take into account stylistic differences.  Specifically, we encoded the documents in the seed set using the biencoder and retrieved summary sentences from the corpus to be used as negatives. Each document summary pair was then scored by the cross-encoder. For a given document, the training objective maximized the score of the correct document summary pair over the incorrect one. The inference stage re-ranking document summary pairs obtained from the biencoder stage using their unnormalized score. 
 
 \section{Downstream Task Model}
 We fine-tune BART on the seed set and the mined data for each task separately.
 \paragraph{Reading Comprehension}
 For SQuAD, we feed the complete document into the encoder and decoder, and use the top hidden state of the decoder as a representation for each word. Start and end classifiers use this representation to classify the token. We simply add the mined data to the initial seed set during fine-tuning.
 
 \paragraph{Summarization}
 For XSum, we use  sequence-to-sequence training with beam decoding during inference. While adding the mined data to the seed set, we denote mined documents using a special token to account for stylistic differences in the documents and summaries.
 
 \section{Experiments}
 \subsection{Benchmarks}
 We perform experiments on SQuAD v1.1 for reading comprehension and XSum for summarization -- two competitive benchmarks which are used to evaluate state of the art pre-trained models. For each dataset, we randomly pick 100 examples from the training set as our seed set; we call this subset $X_{100}$.
 
 \paragraph{SQuAD} SQuAD~\cite{rajpurkar2016squad} is an extractive reading comprehension benchmark containing Wikipedia passages and crowdsourced questions.
 
 \paragraph{Xsum} Xsum~\cite{narayan-etal-2018-dont} is an abstractive news summarization dataset collected by harvesting online
articles from the British Broadcasting Corporation (BBC).

 \subsection{Baselines}
 \paragraph{BART}
  Our main baseline for each dataset is a BART model finetuned on $X_{100}$ and is referred to as BART in the subsequent tables. The fine-tuning differs from our methods (Section~\ref{sec:application}) only in the data used.

  For summarization, we also compare our method to WikiTransfer~\cite{fabbri-etal-2021-improving} and Pegasus~\cite{Zhang2020PEGASUSPW} -- both of which use data augmentation techniques tailored towards summarization.
  
  \paragraph{WikiTransfer} WikiTransfer uses an intermediate finetuning stage with 400K synthetic examples from Wikipedia along with additional auxiliary losses before finetuning on Xsum. 
  
  \paragraph{Pegasus} Pegasus, a self-supervised pretraining method, simulates the summarization tasks during pre-training by masking (and then generating) important heuristically identified sentences from documents. 
 
 \subsection{Implementation Details}
 We implemented all models in fairseq~\cite{ott2019fairseq} and initialized them with BART. The  biencoders were trained for 100 (XSum) or 200 (SQuAD) steps on 8 Volta 32GB GPUs using learning rates (LRs) in the range 5e-6 and \{1,2,3,4,5\}e-5. The crossencoders were trained for 200 (XSum) or 1000 (SQuAD) steps on 8 GPUs using LRs in the same range. We used the accuracy on the dev set for model selection. Both downstream task models are fine-tuned for 200 steps on 8 GPUs using the same learning rates.
 
 \subsection{Main Results}
 \begin{table}
\centering
\begin{tabular}{l c c c}
\toprule
Method & EM & F1 \\
\midrule
BART & 35.27 & 49.43\\
Us & 49.39 & 62.5 \\
\bottomrule
\end{tabular}
\caption{Performance on the SQuAD dev set after training on the seed set of 100 examples, and additionally in case our method, 500 mined examples.}
\label{tab:squad_main}
\end{table}

\begin{table}
\centering
\begin{tabular}{l c c c}
\toprule
Method & R-1 & R-2 & R-L \\
\midrule
BART & 36.28 & 14.07 & 27.98\\
Us & 37.67 & 14.86 & 29.44 \\
BART* & 35.17 & 13.29 & 27.20 \\
WikiTransfer* & 37.26 & 14.20 & 28.85\\
Pegasus* & 39.07 & 16.44 & 31.27 \\
\bottomrule
\end{tabular}
\caption{Performance on the XSum test set after training on 100 examples (and an additonal 500 mined examples for our method). Both BART and Us use the same training examples. * indicates that results have been taken from ~\citet{fabbri-etal-2021-improving} and use a different seed set.}
\label{tab:xsum_test}
\end{table}

\paragraph{SQuAD}
 Table~\ref{tab:squad_main} shows our performance on the SQuAD v1.1 dev set. The BART baseline reaches an F1 of 48.86; adding 500 mined examples gives us a boost of 13 F1 points.   

\paragraph{XSum}
 Table~\ref{tab:xsum_test} shows our performance on the Xsum test set. Augmenting the 100 example training set with 500 mined examples increases the ROUGE-L score of the finetuned BART baseline from 27.98 (BART) to 29.44 (Us).

 BART* refers to BART results taken from WikiTransfer.\footnote{We posit that our numbers for BART are slightly higher due to a different seed set.} Our gains over the BART baseline are comparable to WikiTransfer despite using only 500 augmented examples indicating that our examples are high-quality. It is harder to make more direct controlled comparisons to Pegasus which uses a larger model and a tailored pre-training scheme compared to BART.
 
 Our approach is complementary to both WikiTransfer and Pegasus; in contrast to pre-training (or intermediate fine-tuning) on synthetic or heuristic-collected examples, we focus on mining fewer high quality examples aimed at mirroring the seed set so that they can be used directly during finetuning instead of pre-training.

 \begin{figure}[!t]
\centering
\includegraphics[width=0.5\textwidth, keepaspectratio, clip ]{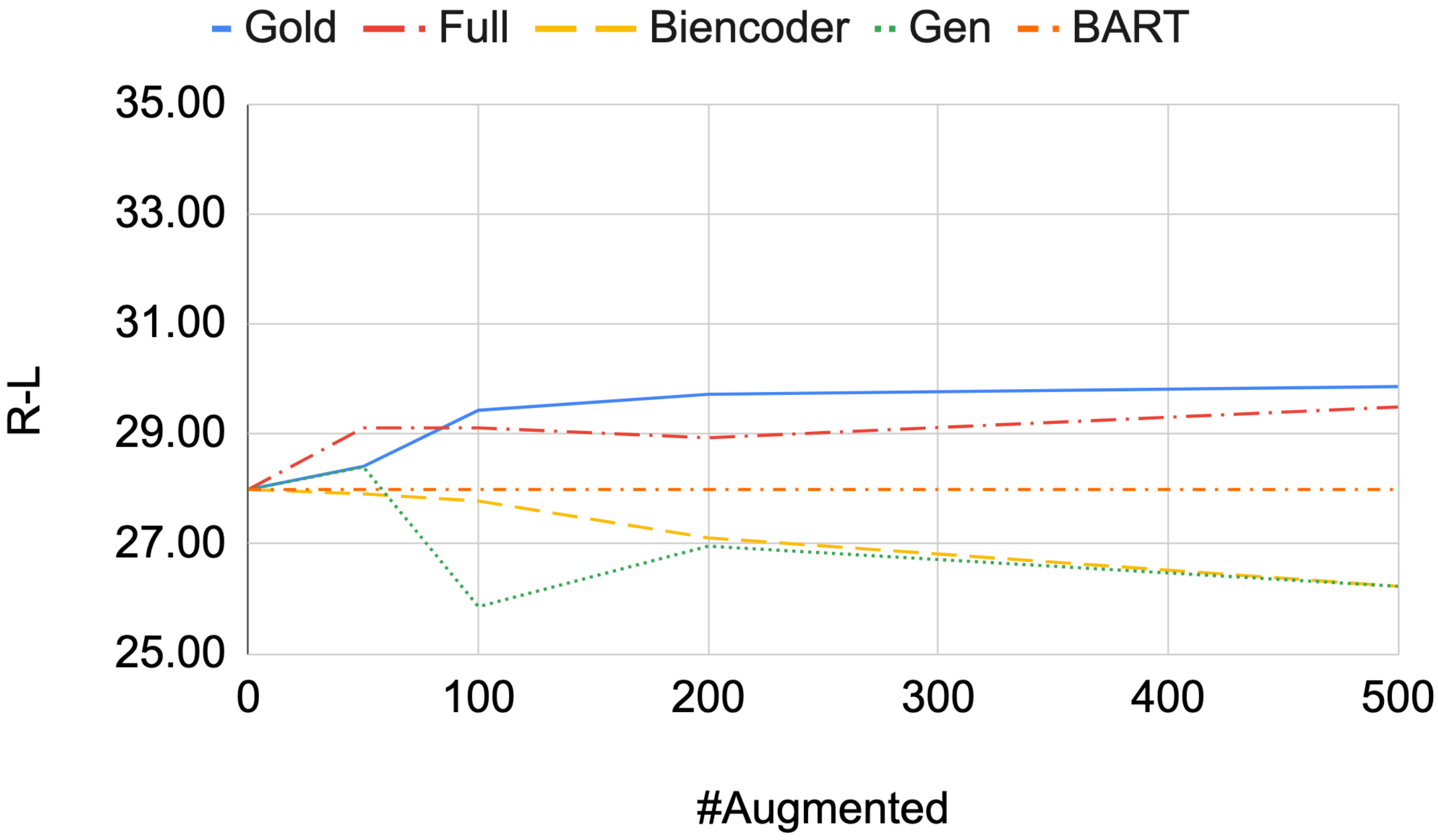}
\caption{Performance (ROUGE-L) on the XSum dev set of various augmentation techniques with varying amounts of augmented data added to $X_{100}$.}
\label{fig:xsum_ablations}
\end{figure}

\begin{figure}[!t]
\centering
\includegraphics[width=0.5\textwidth, keepaspectratio, clip ]{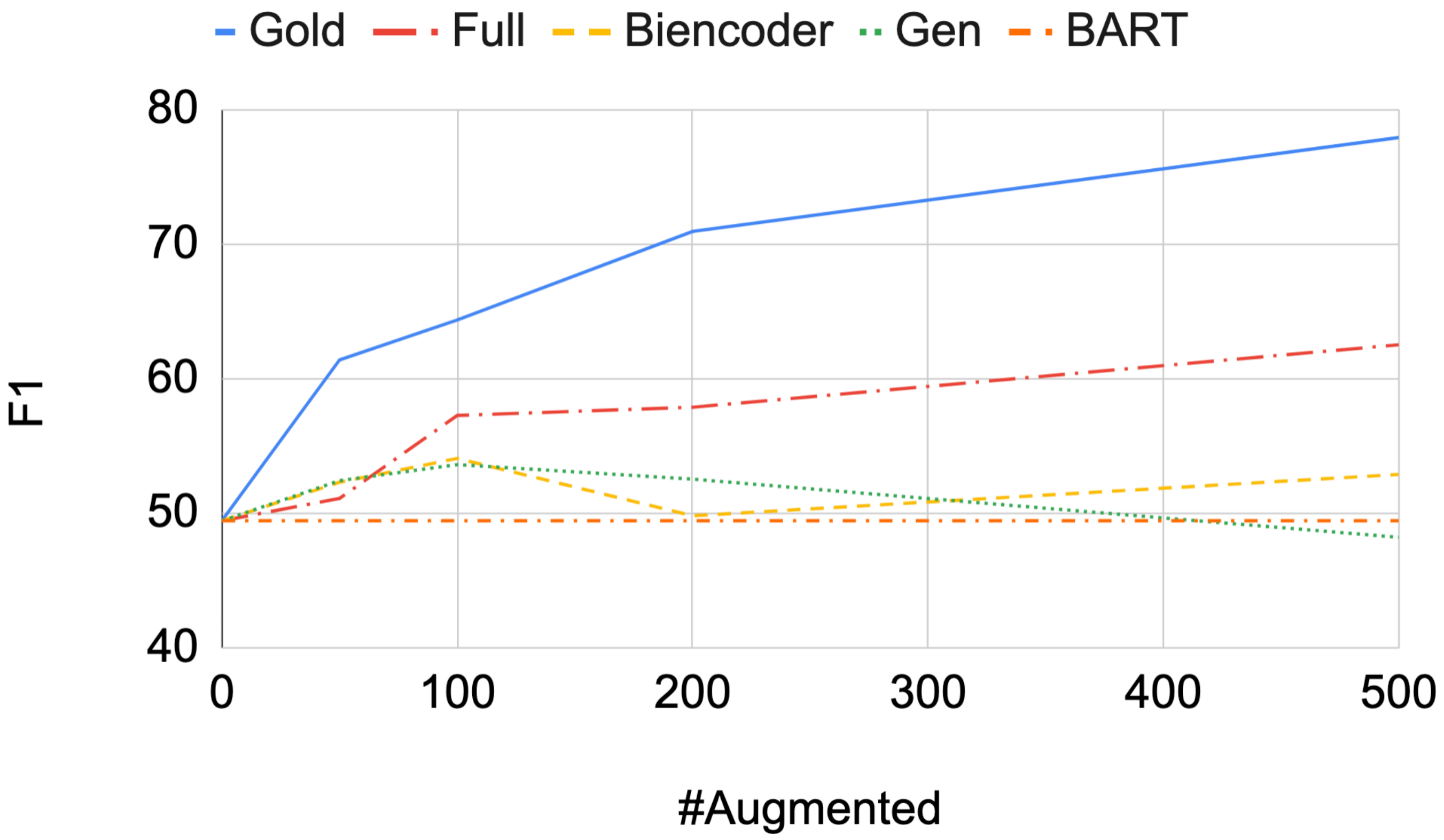}
\caption{Performance (F1) on the SQuAD dev set of various augmentation techniques with varying amounts of augmented data added to $X_{100}$.}
\label{fig:squad_ablations}
\end{figure}
 

 
 \section{Ablations}
 \label{sec:ablations}
 In this section, we investigate the impact of each part of our pipeline and compare it to model-generated data augmentation (Figures~\ref{fig:xsum_ablations} and~\ref{fig:squad_ablations}). All methods involve fine-tuning BART on the same seed set plus the exact same number of augmented examples, but differ on the source of the augmented data.
 
 \paragraph{Gold} The Gold method augments gold examples sampled uniformly at random from the full training set of each dataset.
 \paragraph{Full} Full refers to examples obtained by running our full pipeline including the crossencoder stage.
 \paragraph{Biencoder} This refers to post-processed examples from the biencoder which are subsequently fed to the crossencoder (in our full model).
 \paragraph{Gen} Gen refers to self-training on model generated examples. Specifically, we take source documents (or answer passage pairs in case of reading comprehension) from the full/crossencoder stage and generate summaries (or questions) using the BART baseline trained on $X_{100}$. We then retrain the model by augmenting $X_{100}$ with the generated examples. This allows us to make a controlled comparison to the full model which uses naturally occurring mined examples.
 
\subsection{Amount of Mined Data}
For both SQuAD and XSum, we see an increase in performance  for the Full model as we increase the number of mined examples up to 5x. There is still a considerable difference between our method and Gold indicating that there is still room for improvement in quality. We discuss these qualitative differences in Section~\ref{sec:discussion}.

 \subsection{Bi vs Crossencoder Differences}
 For SQuAD, input output pairs from the biencoder provide some  improvement with 50 examples but performance quickly plateaus with more examples. For XSum, adding examples from the biencoder hurts performance. On the other hand, performance continues to improve until 5x examples from the crossencoder are added for both tasks. Section~\ref{sec:discussion} contains a qualitative analysis of the bi and crossencoder examples.

 \subsection{Naturally Occurring vs. Model Generated Data}
 To show the value of mining naturally occurring data, we compare our method to the \emph{Gen} baseline described in Section~\ref{sec:ablations}. For both XSum and SQuAD, adding model-generated examples provides some gains over BART, but performance quickly degrades with more examples. While model generated data augmentation has been shown to help for QA~\cite{alberti-etal-2019-synthetic}, our results suggest that it isn't particularly useful when training data is limited.
 
 \section{Discussion}
 \label{sec:discussion}
   We conducted a case study of the top ranking examples from the bi and cross encoders for both SQuAD and XSum (Table~\ref{tab:discussion}).

   \begin{table*}[!t]
\centering
\small
\begin{tabular}{ p{.46\linewidth}  p{.46\linewidth}  }
\toprule
Biencoder & Crossencoder \\
\midrule
\textbf{Summary}: BERLIN (AP) — A 44-pound minute hand has fallen off a clock on a Hamburg church tower, plunging 130 feet onto the sidewalk below. & \textbf{Summary}: Blackpool’s three historic piers have been put on an international preservation list amid fears their future could be under threat from climate change.\\

\textbf{Document}: Minute hand plunges from Hamburg church tower; no one hurt BERLIN (AP) — A 20-kilogram (44-pound) minute hand has fallen off a clock on a Hamburg church tower, plunging 40 meters (130 feet) onto the sidewalk below... & \textbf{Document}: ...The council is now completing an action plan to work with the World Monuments Fund which was launched in 1996 with founding partner American Express and issues a list every two years. Blackpool is unique in being the only UK seaside resort with three piers...\\
\midrule
\textbf{Summary}: NEW LONDON, Conn. (AP) — The U.S. Coast Guard Academy is now offering an academic program in cyber systems, its first new major in a quarter century.  & \textbf{Summary}: Holidaymakers are being reminded to drink responsibly at Gatwick Airport as an annual campaign to tackle disruptive passengers is launched. \\

\textbf{Document}: Coast Guard Academy to offer new major in cyber systems NEW LONDON, Conn. (AP) — The U.S. Coast Guard Academy is offering a new major in cyber systems. It's the first new academic program at the school in New London since the addition of mechanical engineering as a major in 1993... & \textbf{Document}: The force, working closely with the airport, its pubs and bars, and airlines, will carry out increased patrols as part of Operation Disrupt..."You could be refused carriage or sent on the next plane home if you are considered to be drunk, disorderly or disruptive...\\
\midrule
\textbf{Question}: What is most times one actor has been nominated? & \textbf{Question}: When
was Regina Lobiondo born?\\
\textbf{Answer/Passage}: The actress has also been nominated \emph{five} times (most & \textbf{Answer/Passage}: Other notable appearances include a recurring role in L.A. Law, a regular role in the 1993 The Untouchables television series, and starring in the 1996 film It's My Party. Regina was born on \emph{October 25, 1956}, in Brooklyn, New York.\\
\midrule
\textbf{Question}: Where was Jesus baptised? & \textbf{Question}: When was Gouverneur Morris born? \\
\textbf{Answer/Passage}: He was baptised at \emph{St Jude's} & \textbf{Answer/Passage}: Gouverneur Morris Jr. Gouverneur Morris II (\emph{February 9, 1813} -- August 20, 1888) was an American railroad executive and the son of a founding father of the United States, Gouverneur Morris. Gouverneur Morris was born on February 9, 1813, Morrisania, Bronx, New York. \\
\bottomrule
\end{tabular}
\caption{Mined examples for summarization and reading comprehension after the bi and crossencoders. Answer spans are indicated in \emph{italics}. Biencoder outputs focus on word overlap leading to (a) abstractive summaries which do not resemble XSum those from XSum in style and topicality, and (b) short noisy passages with with greater proportion of question words for reading comprehension. }
\label{tab:discussion}
\end{table*}

 \subsection{XSum} After examining outputs from the bi and crossencoder, we found that both stages were generally able to match documents to  sentences from the same news story, but with the sentences taken from a parallel document. Exceptions to this included cases where similar events (e.g. World Cup games) occuring over different time frames. However, we found three clear differences which could explain why data from the crossencoder is more helpful. First, summaries from the crossencoder were more abstractive than the biencoder. We measured abstractiveness using ROUGE precision scores with respect to the \emph{source}; higher scores being indicative of higher extractiveness and lower abstractiveness. ROUGE-1/2/L precision scores for the crossencoder summaries (64.91 / 15.42 / 55.59) were closer to the XSum validation (66.85 / 17.94 / 59.82) compared to those from the biencdoer (82.21 / 51.66 / 75.92). Second, the distribution of topics and locations from the top ranked crossencoder documents better matched XSum with most stories centered around UK politics and sports. On the other hand, top ranked pairs from the biencoder were more diverse topically. Lastly, the summaries mined by the biencoder rarely matched the style of XSum and often contained source  markers (e.g. CNN or AP News). Our analysis suggests that the biencoder uses word overlap as it's primary signal for scoring document summary pairs while the crossencoder is able to focus on finer details.  
 
 \subsection{SQuAD}
 We found mining high-precision QA pairs to be a much harder problem in part due to the large search space of over 200M (answer, passage) tuples. Like summarization, biencoder outputs for QA focused on word overlap which resulted in the retrieval of very short passages (see Table~\ref{tab:discussion}). While the crossencoder was more successful in selecting longer passages, we still found pairs with subtle errors. For example, for the question, ``When was Regina Lobiondo born?'', the retrieved passage actually refers to Paul Regina and not Regina Lobiondo. We hypothesize the negative impact of such noise is mitigated in the reading comprehension setting where the given evaluation passage always answer the question. Another limitation of data from both stages was the lack of diversity in question types with a significant proportion of questions seeking date or numerical answers.

 \section{Related Work}
 Perhaps most related to our work is CRISS~\cite{Tran2020CrosslingualRF}, an unsupervised method for mining parallel sentences across languages. Similar to our work, they use a biencoder to mine pairs. However, multilingual pre-training enables the use of cross-lingual cosine similarity functions (over mean pooled output representations) without additional biencoder training. In contrast, for tasks where such high-precision similarity functions aren't readily available, we need (a) a seed set to train biencoders, and (b) a (trained) crossencoder for removing more subtle errors. We divide other related work into the following categories.
 \paragraph{Data augmentation}
 Data augmentation is a popular technique with a large body of work~\cite{wang-yang-2015-thats,jia-liang-2016-data} among others. Recent work has explored model generated data augmentation for a range of tasks including text classification~\cite{Anaby-Tavor_Carmeli_Goldbraich_Kantor_Kour_Shlomov_Tepper_Zwerdling_2020,schick-schutze-2021-generating}, question answering~\cite{alberti-etal-2019-synthetic}, common-sense reasoning~\cite{yang-etal-2020-generative}, and machine translation~\cite{sennrich-etal-2016-improving}. A common problem with model-generated data augmentation is the quality of the synthetic data. Attempts to remedy this have focused either on regularization~\cite{He2020Revisiting} or on variations of consistency~\cite{xie2019unsupervised,sohn2020fixmatch} for a given task--such as round-trip consistency of question generation and answer prediction~\cite{alberti-etal-2019-synthetic,puri-etal-2020-training} for QA or between source and targets in summarization~\cite{fabbri-etal-2021-improving}. Work on retrieval-based data augmentation shows that self-training captures signal complementary to pre-training but  has focused on classification tasks~\cite{du-etal-2021-self}. Ex2~\cite{Lee2021NeuralDA} focus on generating synthetic data for underrepresented or few-shot slices using a model learned by simulating the example generation procedure
on data-rich slices of the data. Task augmentation~\cite{vu-etal-2021-strata} generates data in the target domain by using a model trained on the auxiliary task of natural language inference. In contrast, we focus on a general framework for mining \emph{naturally occurring} data for multiple tasks using supervision from a small labeled seed set.  
 
 \paragraph{Improving fine-tuning}
 Work in this area has looked at improving finetuning particularly for small datasets either via better optimization or by using auxiliary tasks.
 Careful design and hyperparameter choices have a significant impact on performance and stability in limited data settings~\cite{Mosbach2021OnTS,revisit-bert-finetuning}. Likewise, regularization based approaches have also been shown to improve performance~\cite{jiang-etal-2020-smart,aghajanyan2020better}. While we use lessons about careful design choices, our basic models use standard fine-tuning for simplicity. Efforts on the data side have focused on intermediate fine-tuning either by using unlabeled target domain data~\cite{karouzos-etal-2021-udalm,gururangan-etal-2020-dont} or via labeled data from other tasks~\cite{Phang2018SentenceEO,aghajanyan-etal-2021-muppet,vu-etal-2020-exploring}. 

 \paragraph{Few-shot learning}
 GPT-3~\cite{NEURIPS2020_1457c0d6} is perhaps the most prominent recent work in this area which shows that massive language models can perform a variety of tasks if prompted with few input output pairs. Other work has shown that good few-shot performance can also be obtained by combining gradient based optimization with textual prompts~\cite{schick-schutze-2021-just,gao-etal-2021-making,tam-etal-2021-improving,Wang2021EntailmentAF}. Other work on pre-training focuses on tailored masking schemes such as recurrent span masking~\cite{ram-etal-2021-shot} for question answering and salient sentence masking for summarization~\cite{Zhang2020PEGASUSPW}. Our approach is complementary to this line of work in that it is agnostic to pre-training schemes or textual prompts, and focuses instead on supervised automated collection of high-quality data.
 
 \section{Conclusion}
 We presented a method to mine input output pairs from large corpora using supervision from a small seed set of labeled examples. Our approach consisted of two stages -- an efficient cheap biencoder search to identify promising candidate pairs from the corpus and an expensive crossencoder filter for re-ranking the output of the first stage. Our method achieved consistent improvements on XSum and SQuAD, two popular NLP benchmarks, from adding up to 5x examples to the seed set.  Our analysis shows that, compared to model generated data, mined data better matches several characteristics of the gold data (e.g. abstractiveness in summarization). 

\bibliography{eacl2021}
\bibliographystyle{acl_natbib}


\end{document}